\documentclass[journal]{IEEEtran}
\usepackage{graphicx}
\usepackage{subfigure}
\usepackage[numbers,sort&compress]{natbib}

\usepackage{caption2}
\usepackage{amsmath}

\usepackage{cases}
\usepackage{float}
\usepackage{amssymb}
\usepackage{multirow}
\usepackage{mathrsfs}
\usepackage{algorithmic}
\usepackage[ruled]{algorithm2e}
\usepackage{rotating}
\usepackage{balance}
\usepackage{threeparttable}
\usepackage{color}
\usepackage{colortbl}
\definecolor{mygray}{gray}{.9}
\usepackage{makecell}
\usepackage{pdflscape}
\usepackage{siunitx}
\usepackage{stfloats}
\usepackage{verbatim}
\usepackage{setspace}
\usepackage{threeparttable}
\usepackage{supertabular,booktabs}
\usepackage{rotating}
\usepackage{supertabular}
\usepackage{pifont}
\usepackage{xcolor,pict2e}

\usepackage{tikz}
\usepackage{booktabs}
\usepackage{footnote}
\usepackage[pagewise]{lineno}
\usepackage{balance}
\newcommand{\etal}{\textit{et al.}}

\begin{document}
\title{Objective Class-based Micro-Expression Recognition through Simultaneous Action Unit Detection and Feature Aggregation}
\author{Ling Zhou,
        Qirong Mao,~\IEEEmembership{Member,~IEEE,}
        Ming Dong,~\IEEEmembership{Member,~IEEE,}
\thanks{L. Zhou and Q. Mao are with the School of Computer Science and Communication Engineering, Jiangsu University, Zhenjiang, Jiangsu, China.
(e-mail: 2111808003@stmail.ujs.edu.cn, mao\_qr@ujs.edu.cn)}
\thanks{M. Dong is  with the Department of Computer Science, Wayne State University.
(email: ak3389@wayne.edu.)}}

\markboth{}%
{Shell \MakeLowercase{\textit{et al.}}: Bare Demo of IEEEtran.cls for IEEE Journals}

\maketitle
\begin{abstract}
Micro-Expression Recognition (MER) is a challenging task as the subtle changes occur over different action regions of a face. Changes in facial action regions are formed as Action Units (AUs), and AUs in micro-expressions can be seen as the actors in cooperative group activities. In this paper, we propose a novel deep neural network model for objective class-based MER, which simultaneously detects AUs and aggregates AU-level features into micro-expression-level representation through Graph Convolutional Networks (GCN). Specifically, we propose two new strategies in our AU detection module for more effective AU feature learning: the attention mechanism and the balanced detection loss function. With those two strategies, features are learned for all the AUs in a unified model, eliminating the error-prune landmark detection process and tedious separate training for each AU. Moreover, our model incorporates a tailored objective class-based AU knowledge-graph, which facilitates the GCN to aggregate the AU-level features into a micro-expression-level feature representation. Extensive experiments on two tasks in MEGC 2018 show that our approach significantly outperforms the current state-of-the-arts in MER. Additionally, we also report our single model-based micro-expression AU detection results.
\end{abstract}


\IEEEpeerreviewmaketitle
\section{Introduction}
\begin{figure}
  \centering
  \includegraphics[width=0.45\textwidth]{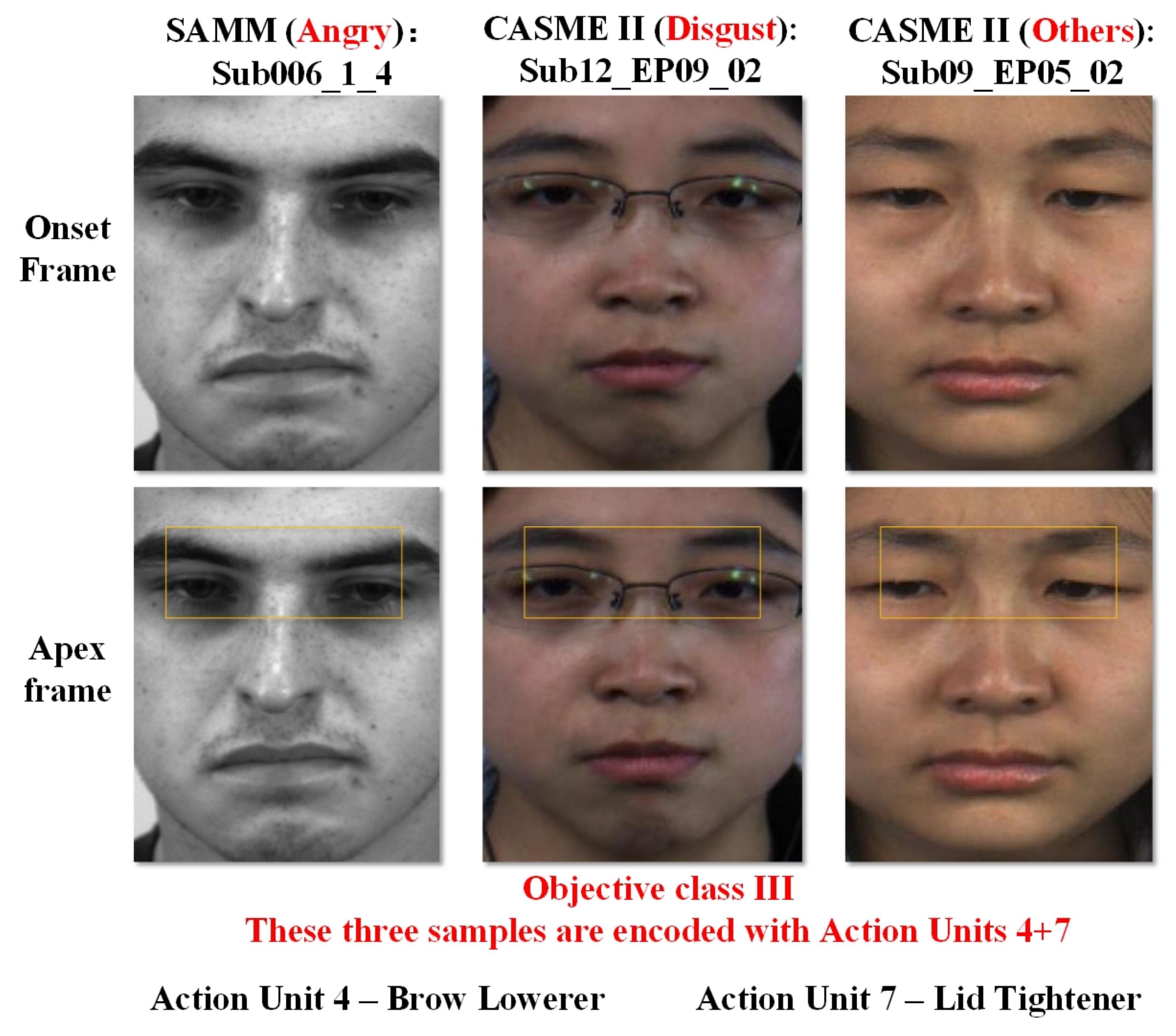}
  \caption{
 Sample frames showing objective class III encoded with the same action units but different emotion labels from SAMM \cite{Davison2018} and CASME II \cite{Yan2014} databases. To catch the movement of the expressions more intuitively, only onset and apex frames are illustrated. Onset and apex frames display neutral expression and the highest intensity changes among all frames of each sample, respectively. The left to right columns show three samples labeled to ``Anger", ``Disgust" and ``Others" emotional classes while encoded with the same action units of ``$4+7$". This inconsistence is the reason why researchers added further  justification for the introduction of new objective classes based on action units only \cite{Davison2018a}.
  }\label{demo}
\end{figure}

Micro-expressions, different from macro-expressions or subtle expressions, are hidden emotions and occur over small regions of a face. Micro-Expression Recognition (MER) has a tremendous impact on a wide range of applications including psychology, medicine, and police case diagnosis \cite{Ekman1969,  Ekm2009}.
The aim of MER is to reveal the hidden emotions \cite{Michael2010,  Ekm2009} of humans and help to understand people's deceitful behaviors when micro-expressions occur. In general, there are two well-known and well-accepted types of MER: emotional-based MER and objective class-based MER \cite{Davison2018a}. Emotional-based MER deals with the recognition on emotion labels, while objective class-based MER tackles the category task where samples are labeled based on the Facial Action Coding System (FACS) \cite{Ekman1978}.

In the emotional-based MER, it was found that there are many conflicts in the coded action units (AUs) and the estimated emotions, mainly due to the unpredictability and bias brought by self-reports \cite{Davison2018a, Yap2018}. One such example is illustrated in Figure \ref{demo}, where three samples of the objective class III from CASME II \cite{Yan2014} and SAMM \cite{Davison2018} are encoded with the same action units ($4+7$) but categorized to different emotion classes. On the other hand, objective classes defined based on AUs can alleviate those conflicts and prompt a more objective MER \cite{Davison2018a}. Classifying micro-expressions on objective labels, instead of predicted emotions, can remove the potential bias of human reporting and thus improve micro-expression recognition \cite{Davison2018a}. Recently, the AU-centric objective classes are proposed for a fair categorization based on FACS \cite{Ekman1978}. The objective classes I-V are linked with happiness, surprise, anger, disgust, and sadness, but not directly correlate to being these emotions \cite{Davison2018a}. The criterion of establishing these links is based on previous psychological research \cite{Ekman1978,Ekman1978a}.

To promote research on objective class-based MER, a single composite database with objective class labels and two related MER tasks are recently introduced in Micro-Expression Grand Challenge (MEGC) 2018 \cite{Yap2018}. Three handcrafted features commonly used in MER has been implemented by Davison et al. \cite{Davison2018a} as the baselines of MEGC 2018, including the popular appearance-based feature Local Binary Pattern from Three Orthogonal Planes (LBP-TOP) \cite{Zhao2007}, 3D Histograms of Oriented Gradients (3DHOG) \cite{Polikovsky2009}, and a geometric-based feature Histogram of Oriented Optical Flow (HOOF) \cite{Liu2016}. However, it is unclear if these handcrafted features by traditional methods can sufficiently characterize the AU-centric objective classes. More recently, Deep Neural Networks (DNN) and optical flow features have been introduced to learn data-driven features for objective class-based MER and achieved state-of-the-art results \cite{Peng2018, Khor2018}.

Fusing local features to a global representation, such as Bag of Words \cite{Csurka2004} and region feature-based deep learning, has shown very promising results in image understanding \cite{Yang2009, Wang2020}. Although new advances have been made in objective class-based MER, they all focused on global features - learning features directly from a whole face image. How to learn local features (e.g., at the AU-level) and aggregate them to a global, micro-expression-level, a very important aspect of MER, is not addressed by current approaches. Recent work on AU-based local feature extraction achieved outstanding performance in AU recognition \cite{Li2017, Li2019a}, but they typically require bounding boxes of AU regions as the input and thus are heavily influenced by the performance of facial landmark detection. The interaction of AUs has also been explored using graph-based networks and was shown to be helpful for AU recognition \cite{Liu2020, Li2019a, Shao2020}.

In general, the micro-expressions can be considered as cooperative group activities which are acted by different AUs. In this paper, we propose a novel deep learning model, MER-auGCN, to perform objective class-based MER through simultaneous AU detection and AU-level to micro-expression-level feature aggregation. Specifically, MER-auGCN first learns all local AU features in a unified model, eliminating the error-prune landmark detection process and tedious separate training for each AU. Then, it aggregates AU-level features into the micro-expression-level feature through
Graph Convolutional Networks (GCN) \cite{Kipf2017}. With the guidance of a tailored objective class-based AU knowledge-graph, a GCN is employed for updating the relational feature of each AU by aggregating from its neighbor nodes. Simultaneous AU detection and feature aggregation are implemented in our MER-auGCN system with three major components: (1) a backbone to extract the global facial features, (2) a unified AU detection module to learn all AU features in one-shot, and (3) a GCN-based feature aggregation module with a final MER classifier.
The major contribution of our work is summarized as follows.
\begin{itemize}
\item To the best of our knowledge, this is the first work that performs simultaneous action unit detection and feature aggregation for objective class-based MER. In MER-auGCN, AU-level (local) and micro-expression-level (global) features are learned collaboratively through AU detection and AU feature aggregation, leading to a more streamlined and accurate MER.

\item By introducing the attention mechanism and the penalization of the balanced AU detection loss, AU features are learned in a single model without any priori facial landmarks, eliminating the error-prune landmark detection process and tedious separate model training for each AU.

\item MER-auGCN models AU interactions with a tailored knowledge-graph, built on the AU correlations in the objective class definition and the training datasets, and learns an effective micro-expression-level feature representation using GCN.

\item The proposed approach achieves state-of-the-art results on two tasks of MEGC 2018: the Holdout-database Evaluation (HDE) task and Composite database evaluation (CDE) task. Micro-expression AU detection results also demonstrate that our model can learn effective AU features.

\end{itemize}
The rest of the paper is organized as follows. Section II introduces the related works. Section III presents our proposed algorithm in detail. Section IV describes the MER benchmark datasets and reports our experimental results. Conclusions are discussed in Section V.

\section{Related Work}
In this section,  we first review the previous work related the feature extraction for objective class-based MER, including both handcrafted features and deeply learned features. As objective classes are defined based on AUs, we also briefly review AU detection for micro-expressions and facial expression recognition.

\subsection{Features in objective class-based MER}
\textbf{Handcrafted features}
\textcolor{black}{
The success of existing~traditional approaches in MER is attributed in good part to the quality of the handcrafted visual features representation. The handcrafted features are generally categorized into appearance-based and geometric-based features. Most handcrafted features are originally designed for emotional-based MER. Three of them were implemented in \cite{Davison2018a} as the baselines for objective class-based MER, including Local Binary Pattern-Three Orthogonal Planes (LBP-TOP) \cite{Zhao2007}, 3D Histograms of Oriented Gradients (3DHOG) \cite{Polikovsky2009, Polikovsky2013} and Histograms of Oriented Optical Flow (HOOF) \cite{Chaudhry2009, Liu2016}. LBP-TOP is the most widely used appearance-based feature for MER. It combines the temporal features along with the spatial features from three orthogonal planes of the image sequence. Due to its low computational complexity, many LBP-TOP variants have been proposed \cite{Wang2014, Wang2015, Huang2015, Huang2016, Huang2019, Zong2018a}. Besides LBP-TOP, 3DHOG is another appearance-based feature that is obtained by counting occurrences of gradient orientation in localized portions of a given image sequence.
Different from LBP-TOP and 3DHOG which are based on the image appearance, Main Directional Mean Optical Flow (MDMO) \cite{Liu2016} aims to represent the micro-expression samples by the aspect of face geometry, e.g. shapes and location of facial landmarks. As a variant of the HOOF feature, MDMO is a low-dimensional geometric-based feature that reduces the feature dimension by using the optical flow in the main direction, thus it is least affected by the varied number of frames in the image sequence.}

\textbf{Deeply learned features}
\textcolor{black}{
Deep learning has been considered as a more efficient way to learn feature representations for MER. Several deep learning methods have been proposed for emotional-based MER, such as Shallow Triple Stream Three-dimensional CNN (STSTNet) \cite{Liong2019}, Dual-Inception \cite{Zhou2019}, Spatiotemporal Recurrent Convolutional
Networks (STRCN)\cite{Xia2020}, etc. Recently, two deeply learned features were presented for objective class-based MER by Peng et al. \cite{Peng2018} and Khor et al.~\cite{Khor2018}. Specifically, to better represent the subtle changes in micro-expression, Khor~\etal~\cite{Khor2018} adopted Enriched Long-term Recurrent Convolutional Network (ELRCN) based on optical flow features. It contained the channel-wise for spatial enrichment and the feature-wise for temporal enrichment predicted the micro-expression by passing the feature vector through LSTM. Peng et al. \cite{Peng2018} adopted pre-trained ResNet10 \cite{Simon2016} as a backbone and introduced a transfer learning strategy to improve MER performance. The RestNet10 was first trained on ImageNet \cite{Russakovsky2015}, then fine-tuned on four public macro-expression databases, finally fine-tuned on the CASME II and SAMM databases by using apex frames. With the delicate pre-training process and transfer learning strategy on the macro-expression databases, \cite{Peng2018} achieved the best performance on objective class-based MER in MEGC 2018 \cite{Yap2018}.}

However, these aforementioned methods all focused on learning global facial features and ignored the AU correlations, which is critical for objective class-based MER. Our work differs from these approaches as it explicitly learns local AU features through AU detection and generates global micro-expression-level feature representation via an AU knowledge-graph guided feature aggregation.

\textcolor{black}{
\subsection{Action Unit Detection}
Due to the limited representation capability of handcrafted feature in AU detection \cite{Eleftheriadis2015,Zhao2016,Song2015}, deep networks \cite{Li2019, Li2017,Li2017a,Kollias2019,Liu2020,Li2019a} have been recently introduced and made unprecedented progress.
For AU detection in micro-expressions, Li et al. \cite{Li2019} proposed Spatio-Temporal Adaptive Pooling (STAP) network to learn multi-scale information on spatial and temporal domains with less computational cost by aggregating a series of convolutional filters of different sizes. For AU detection in facial expressions, Li et al. \cite{Li2017} presented an adaptive region cropping based multi-label learning deep recurrent network to acquire global feature across sub-regions by multi-label learning. Specifically, the features of the sub-regions are learned separately in region of interest (ROI) cropping nets. Meanwhile, EAC-Net \cite{Li2017a} was established to conduct enhanced feature learning for facial regions with a relatively large local region to cover the AU target areas and achieved good performances. More recently, to learn semantic features for AU detection, some researchers attempted to explore the semantic relationship between AUs to improve the AU detection performance. Liu et al. \cite{Li2019a} used Gated Graph Neural Network (GGNN) \cite{Li2016}  to learn robust features in the semantic relationship embedded representation learning (SRERL) framework  by analyzing the symbiosis and mutual exclusion of AUs in various facial expressions. Different from SRERL, Liu et al. \cite{Liu2020} built the AU knowledge-graph according to the confidence of AUs relations in the training database, and leveraged GCN \cite{Kipf2017} for AU relational feature learning.}

\textcolor{black}{However, these AU detection methods mentioned above either need training different models for different AUs or learned individual AU features with ROI, which
are heavily influenced by the performance of facial landmark detection.  In our work, we introduced AU detection as an intermediate step for AU feature learning, as our final goal is MER using the learned features. Moreover, AU detection in our system addressed the limitations of existing AU detection approaches as it learned features of AUs in an attention-based unified model, eliminating the error-prune landmark detection process and tedious separate training for each AU.}

\begin{figure*}
  \centering
  \includegraphics[width=\textwidth]{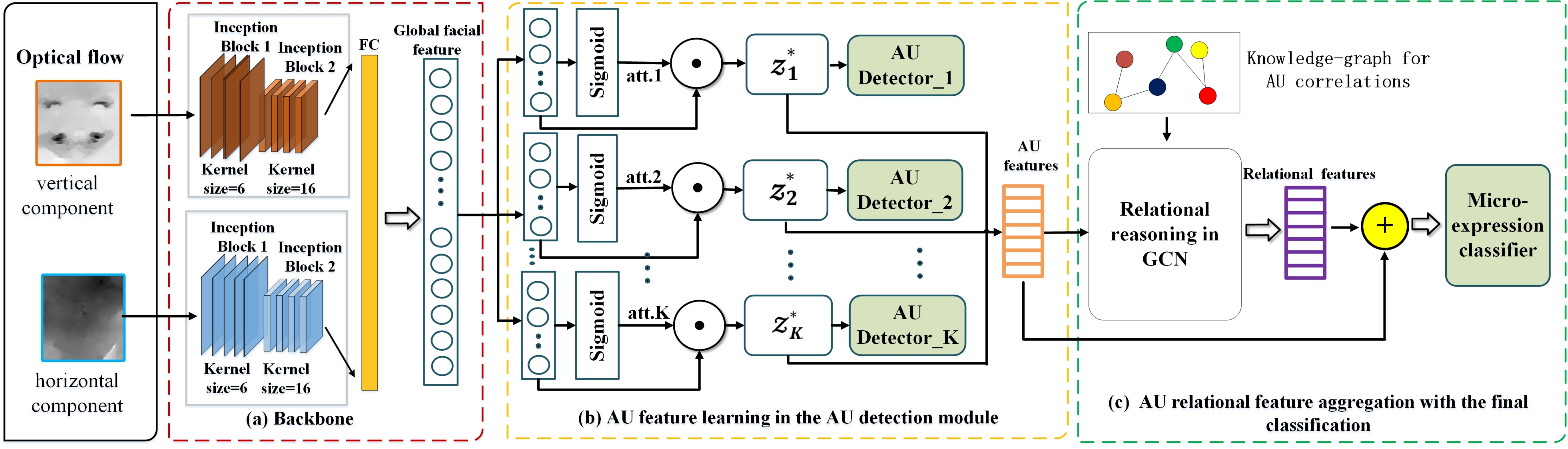}
  \caption{
  The architecture of the proposed MER-auGCN model for objective class-based MER. We first feed optical flow images into the backbone to extract the global facial feature of each sample. Then we learn individual AU features simultaneously through $\textbf{\emph{Sigmoid}}$ attention based AU detection module. Afterwards, a tailored objective class-based AU knowledge-graph is incorporated, which facilitates the GCN to aggregate the AU-level features for final classification.}\label{fr}
\end{figure*}
\section{The MER-auGCN model}
The micro-expressions can be considered as cooperative group activities which are acted by different AUs. In MER-auGCN, we first extract features for each AU through AU detection, and then aggregate them into the micro-expression-level via knowledge-graph guided GCN. In this section, we first introduce the construction of the AU relational graph, which works as a guided knowledge-graph for GCN-based feature aggregation in our framework. Then, we present the details of our MER-auGCN model as illustrated in Figure \ref{fr}.

\begin{table}
\caption{Relationship between action unit and objective classes I-V.}
\small
\label{objective_class}
\begin{center}
\setlength{\tabcolsep}{0.4mm}{
\begin{tabular}{|c|l|}
\hline
Class & Action Units \\
\hline
I&AU6, AU12, AU6+AU12, AU6+AU7+AU12, AU7+AU12\\
\hline
\multirow{2}*{II}&AU1+AU2, AU5, AU25, AU1+AU2+AU25, AU25+AU26,\\
&AU5+AU24\\
\hline
\multirow{2}*{III}&A23, AU4, AU4+AU7, AU4+AU5, AU4+AU5+AU7,\\
&AU17+AU24, AU4+AU6+AU7, AU4+AU38\\
\hline
\multirow{3}*{IV}&AU10, AU9, AU4+AU9, AU4+AU40, AU4+AU5+AU40,\\
&AU4+AU7+AU9, AU4 +AU9+AU17, AU4+AU7+AU10,\\
&AU4+AU5+AU7+AU9, AU7+AU10\\
\hline
V&AU1, AU15, AU1+AU4, AU6+AU15, AU15+AU17\\
\hline
\end{tabular}}
\end{center}
\end{table}

\begin{figure}
  \centering
  \includegraphics[width=0.48\textwidth]{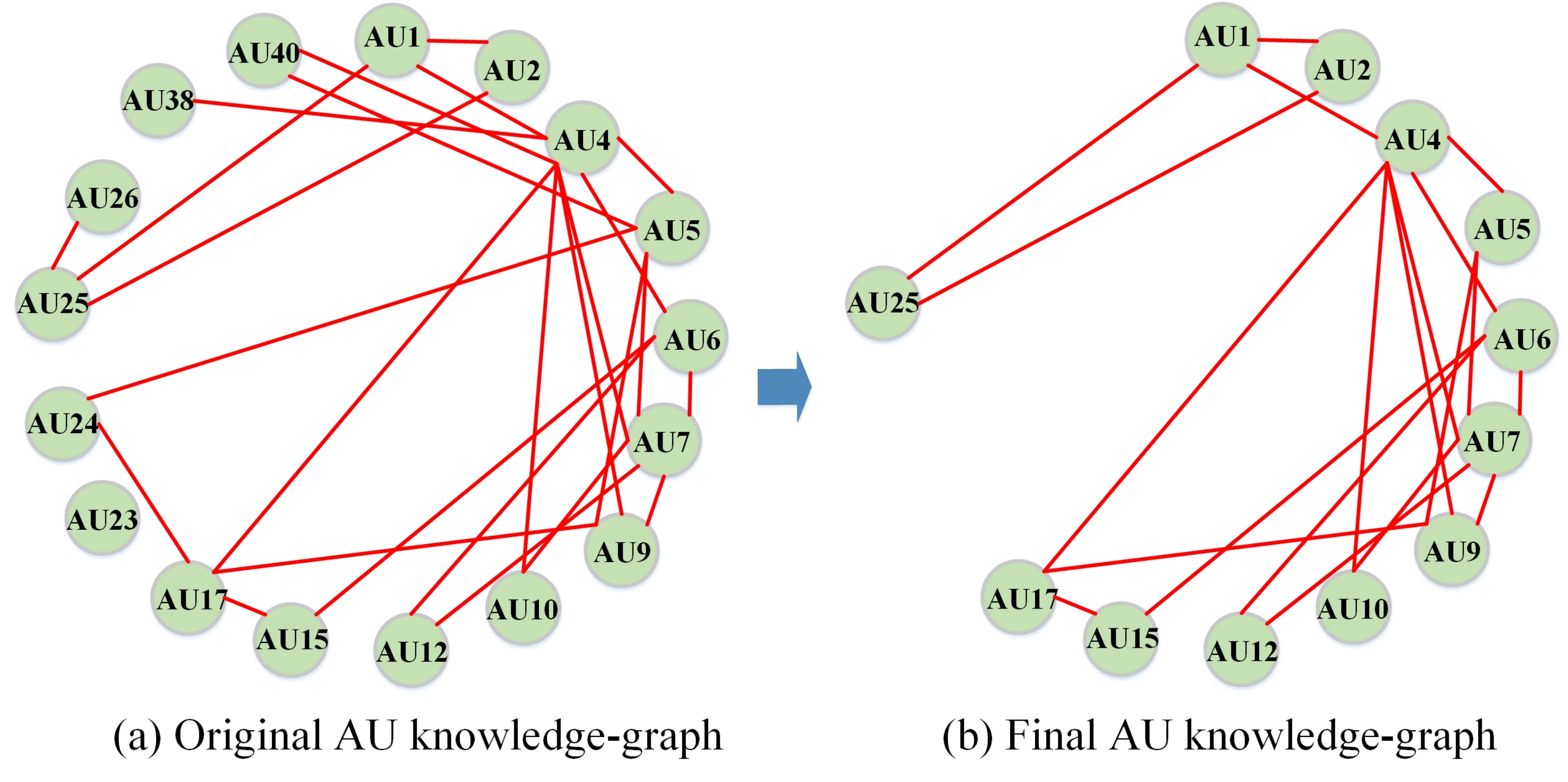}
  \caption{
  The process of constructing the AU knowledge-graph. We first model the original knowledge-graph according to the AU correlations of the objective class definition, then adjust it referring to the training databases. Each node in the graph is connected to itself. For demonstrating the correlations between different AUs clearly, the node-self link edges are not depicted.
  }\label{graph}
\end{figure}

\subsection{Construction of AU knowledge-graph}
The aim of constructing the AU relational graph in our work is to establish the semantic relationship between AUs and the objective classes. The single composite database for MEGC 2018 tasks \cite{Yap2018} consists of samples from SAMM \cite{Davison2018} and CASME II \cite{Yan2014} databases and has five objective classes. Table \ref{objective_class} shows the five objective classes (I-V) and the corresponding AUs that have been assigned to each class \cite{Davison2018a}, where AUs linked with ``$+$" indicate that these AUs could co-exist in one micro-expression class. 

As shown in the left panel of Figure \ref{graph}, the original AU knowledge-graph is built following relations defined in Table \ref{objective_class}. It contains a total of 17 nodes, corresponding to the 17 AUs. Edges in the graph mean the two linked AUs can co-exist in the same micro-expression sample. The HDE task in MEGC 2018 includes two sub-tasks: one is training on the CASME II database and testing on the SAMM database (\emph{CASEM II $\rightarrow$ SAMM}), and the other is training on SAMM and testing on CASME II (\emph{SAMM $\rightarrow$ CASME II}). In the sub-task of \emph{CASEM II $\rightarrow$ SAMM}, the AUs that do not exist in CASME II (i.e., AU23, AU24, AU26 and AU40) will bring additional noises to the relational reasoning because their features cannot be learned without training data. The same holds for the sub-task of \emph{SAMM $\rightarrow$ CASME II}, where AU38 and AU40 do not exist in the training set of SAMM. Thus, we removed those AUs (AU23, AU24, AU26, AU38, and AU40) from the original relational graph and got an undirected symmetric AU knowledge-graph $\mathbf{G}$ with $12$ nodes shown in the right panel of Figure \ref{graph}. If there is a edge between the node $i$ and $j$, $\mathbf{G}_{i,j} = 1$, and otherwise $\mathbf{G}_{i,j} = 0$. Note that in addition to the edges between different AUs, each node in the graph is connected to itself even though these edges are not depicted to show a cleaner relationship between different AUs.

\subsection{Feature learning in MER-auGCN}
As shown in Figure  \ref{fr}, our proposed MER-auGCN framework consists of three key components: the global facial feature extraction with the backbone, the AU feature learning in the AU detection branch, and the AU relational feature aggregation with the final classification. In the following, we present the three modules in details.

\subsubsection{Global facial feature extraction}
We choose the Dual-Inception \cite{Zhou2019} model as our backbone network to learn global facial features.  Dual-Inception is composed of two groups of Inception \cite{Szegedy2015} blocks with two layers of depth. This shallow network has been used for MER using optical flow images as the input. Besides that, we also have conducted an experiment on another deep-depth network backbone model to verify the effectiveness of our approach for MER. As shown in the part (a) of Figure \ref{fr}, after the processing of face cropping and optical flow calculated from the onset and apex frames of each sample, we feed the two components of the optical flow into the two paths of the backbone. Afterwards, we concatenate the two feature maps learned by the two-path inception blocks. Finally, a fully connected layer is performed on the concatenated feature to get a $d$-dimensional global facial feature for each sample, where $d = 1024$ in our model.

\subsubsection{AU feature learning}
AU feature learning for AU detection is mainly based on the landmark-based bounding boxes for AUs, and often requires separate detector training for each AU.  To eliminate the error-prune landmark detection process and learn features of all the AUs simultaneously, a $Sigmoid$ attention mechanism and a balanced AU detection loss are introduced in MER-auGCN.

\textbf{$\textbf{Sigmoid}$ attention:} The attention mechanism provides the flexibility to our model by avoiding the detection of landmarks as the preprocess step to learn AU features. $K$ attention layers are shown in the part (b) of Figure \ref{fr}, and a $d$ dimensional global facial feature $z$ is connected with $K$ fully connected layers separately. After activated by $Sigmoid$, we gain the attention weight of the feature for each AU. Then, representation $ z_k^* $ for each AU is given by:
\begin{equation}
z_k^*= {Sigmoid}_k(z)\ast z,
\label{Sigmoid}
\end{equation}
where $k\leq K$, and $K$ is the number of AUs. ${Sigmoid}_k(z)$ is the attention weights of the factors for the $k$-th AU gained after activated $Sigmoid$ function.

\textbf{Balanced detection loss:}
In addition to the $Sigmoid$ attention, our AU feature learning is also constrained by a set of AU detection loss in the AU detectors. Each AU detector contains a multi-layer perceptron with an $Sigmoid$ layer as the output and is aligned with the feature vector $z_k^*$ as the input. Similar to the AU detection task \cite{Li2019a, Li2017}, the occurrence rate of AUs in MER is heavily imbalanced. To better tackle the AU feature learning task with imbalanced data, we design an adaptive function with the focal loss \cite{Lin2017a} to balance the proportion of the negative and positive samples in a given training batch. Specifically, we first define a balance factor $\alpha$ adapted to each training batch, which is computed as the proportion of the positive samples in a training batch,
\begin{equation}
\begin{split}
\alpha_i = \frac{\sum_{j=1}^M[y_{i,j}^{D}=1]}{M},
\end{split}
\label{alfa}
\end{equation}
where $M$ denotes the training batch size, and $y_{i,j}^{D}$ is the binary label of the $j$-th sample in a training batch belonging to the $i$-th AU, either 1 or 0. Then, all AU detectors are trained simultaneously by optimizing the following adaptive focal loss,
\begin{equation}
\begin{split}
\mathcal{L}_{detc}(\theta;y^{D})=-\frac{1}{K\cdot M}\sum_{j=1}^{M}\sum_{i=1}^{K}[\alpha_i\cdot{(1-p(y_{i,j}^{D}))}^2\cdot\\
y_{i,j}^{D}\log(\frac{p(y_{i,j}^{D})+0.05}{1+0.05}) +  (1-\alpha_i)\cdot p(y_{i,j}^{D})^2\cdot\\
(1-y_{i,j}^{D})\log(\frac{1+0.05-p(y_{i,j}^{D})}{1+0.05})],
\end{split}
\label{auloss}
\end{equation}
where $K$ denotes the number of AUs, and $p({y}_{i,j}^{D})$ is the probability of the $j$-th sample detected as the $i$-th AU. The offset of 0.05 in the loss is to prevent extreme loss explode which will stop the training \cite{Li2017}.

With the attention mechanism and the balanced detection loss in each training batch, the features associated with the $K$ individual AUs are obtained simultaneously from the global facial feature. The dimension of each AU feature is the same as that of the global facial feature, and thus we use a $K \times d$ matrix $\mathbf{X}$ to represent the feature vectors of the $K$ AUs.

\subsubsection{AU relational feature aggregation with GCN}
Referring to the part (c) of Figure 1, in this section, we give a detailed description on how to aggregate the AU-level features into a micro-expression-level feature representation using GCN.

Using the matrix $\mathbf{X}$ of AU features and the AU knowledge-graph $\mathbf{G}$ as the inputs, we can perform relational reasoning with GCN to facilitate a more accurate MER. The graph convolutions in GCN allow each target node in the graph to aggregate features from all neighbor nodes according to the edge weight between them. As messages can be passed inside the graph to update features on each node, the output of the GCN provides an updated relational features of each detected AU in the input sample. Formally, one layer of the GCN can be represented as,
\begin{equation}
\begin{split}
\mathbf{Z}^{(l+1)} = \sigma (\mathbf{G}\mathbf{Z}^{(l)}\mathbf{W}^{(l)})\\
\end{split}
\label{gcn}
\end{equation}
where $\sigma(.)$ denotes the ReLU activation function, $\mathbf{G} \in \mathbb{R}^{K\times K}$ denotes the matrix representation of the AU knowledge-graph, $\mathbf{Z}^{(l)}\in \mathbb{R}^{K\times d_l}$ is the AU feature representation in the $l$-th graph convolution layer with $\mathbf{Z}^{(0)} = \mathbf{X}$, and $\mathbf{W}^{(l)}$ is the layer-specific learnable weight matrix. Multi-layer can be stacked into this layer-wise propagation, and we used two layers of GCN in our work. The feature dimension of each AU after the second GCN layer $d_2$ is set at 64, and AU relational features are sumpooled together to generate a 64 dimension micro-expression-level feature vector $z_r$ for each sample.

Finally, we fuse the micro-expression-level feature representation with the AU features and feed it into a $Softmax$ classifier. First, the AU features in $\mathbf{X}$ are sumpooled to $z_o$. Then, the $z_o$ is adjusted to the same dimension as $z_r$ through a fully connected layer. Afterwards, the two representations $z_o$ and $z_r$ are fused via the element-wise summation and used for the final MER by minimizing the standard cross-entropy loss,
\begin{equation}
\mathcal{L}_{cls}(\phi;y^C)=-\sum_{i=1}^{N}[y_i^C\log(\frac{exp(F_i)}{\sum_i {exp(F_i)}})],  \\
\label{cls}
\end{equation}
where $N$ is the total training samples, $y_i^C$ is the objective class label for the $i$-th training instance, and $F_i$ denotes the last fully connected layer in the model which is activated by a $Softmax$ unit. Besides the element-wise summation, we also evaluated concatenation as a fusion function in our experiments.

Combining the balanced detection loss with the standard cross-entropy loss, the final loss function is give as:
 \begin{equation}
\mathcal{L}(\theta, \phi;y^C,  y^D)= \mathcal{L}_{cls}(\phi;y^C) + \lambda\mathcal{L}_{detc}(\theta;y^D)
\label{totalloss}
\end{equation}
where the hyper-parameter $\lambda$ is used to balance the AU detection error and micro-expression classification error. By optimizing the joint loss function, our MER-auGCN model can learn AU-level (local) and micro-expression-level (global) features collaboratively.

\section{Experiments}
\subsection{Datasets}
In MEGC 2018, two spontaneous micro-expression benchmark datasets, CASME II \cite{Yan2014} and SAMM \cite{Davison2018}, are combined into a single database. Even though there are seven objective classes in total, the focus of MEGC 2018 is on the first five: objective classes I to V, which contain 185 samples of 26 subjects in CASME II and 68 samples of 21 subjects in SAMM, respectively. Table \ref{data} provides a summary of the composite database, which has a total of 253 micro-expressions categorized into objective classes I to V.
\begin{table}
\caption{Sample information of objective classes I-V in CASME II and SAMM datasets.}
\label{data}
\begin{center}
\setlength{\tabcolsep}{1.3mm}{
\begin{tabular}{|m{2.3cm}<{\centering}|c|c|c|c|c|c|c|}
\hline
\multirow{2}*{Database}&\multicolumn{6}{c|}{Objective Class}&\multirow{2}*{Subjects}\\
\cline{2-7}
&I&II&III&IV&V&Total&\\
\hline
CASME II \cite{Yan2014}&25&15&99&26&20&185&26\\
\hline
SAMM \cite{Davison2018}&24&13&20&8&3&68&21 \\
\hline
Composite&49&28&119&34&23&253&47\\
\hline
\end{tabular}}
\end{center}
\end{table}
\subsection{Experiment setup and implementation details}
\subsubsection{Experiment setup}

In our experiment, we evaluated the MER-auGCN model on three tasks: two micro-expression recognition tasks (the HDE and CDE tasks) in MEGC 2018 \cite{Yap2018} and a micro-expression AU detection task.

\textbf{HDE task:} There is a 2-fold cross-validation in the HDE task, i.e., training on CASME II and testing on SAMM (\emph{CASME II$\rightarrow$SAMM}), vice versa (\emph{SAMM$\rightarrow$CASME II}). Following the setup in MEGC 2018, unweighted average recall (UAR), weighted average recall (WAR) \cite{Schuller2010}, and the averaged WAR and UAR on both folds were used to measure the performance of different approaches.

\textbf{CDE task:} The CDE task is evaluated with Leave-One-Subject-Out (LOSO) cross-validation. In this task, all the samples from CASME II and SAMM are combined into a single composite database. Samples of each subject are held out as the testing set while all the remaining samples are used for training. F1 score and Weighted F1 score (WF1) were employed to measure the performance of various methods for the CDE task. Here, the F1 is an average of the class-specific F1 across the whole classes (or macro-averaging), and WF1 is weighted by the number of samples in corresponding classes before averaging \cite{Yap2018}.

Specifically, the four metrics (WAR, UAR, F1, WF1) mentioned above are calculated as follows:
 \begin{equation}
WAR = \frac{\sum_{c=1}^{C}TP_c}{N},
\label{war}
\end{equation}
\begin{equation}
UAR = \frac{1}{C}\sum_{c=1}^{C}\frac{TP_c}{N_c},
\label{uar}
\end{equation}
\begin{equation}
F1 = \frac{1}{C}\sum_{c=1}^{C}\frac{2\cdot TP_c}{2\cdot TP_c + FP_c+FN_c},
\label{f1}
\end{equation}
\begin{equation}
WF1 = \sum_{c=1}^{C}\frac{N_c}{N}\frac{2\cdot TP_c}{2\cdot TP_c + FP_c+FN_c},
\label{f1}
\end{equation}
 where $C$ is the number of classes, $c\leq C$, $N_c$ is the number of samples in the ground truth of the $c$-th class, and $N$ is the total samples. $TP_c$,  $FP_c$,  and $FN_c$ are the  true positives, false positives and false negatives of the $c$-th class, respectively.

\textbf{AU detection task:} In addition to the HDE and CDE tasks in MEGC 2018, we also conducted a micro-expression AU detection experiment to verify the effectiveness of our model on AU feature learning. We used the F1 score to measure the performance of various micro-expression AU detection methods. The F1 for each AU is $\frac{2\cdot TP}{2\cdot TP + FP+FN}$, where $TP$ is the true positives, $FP$ is the false positives, and $FN$ is the false negatives.

\subsubsection{Implementation details}
In the pre-processing step, we utilized the algorithm of libfacedetection \cite{Yu2016} to crop out the facial area in each sample and then extracted TV-L1 optical flow \cite{Zach2007} images from the onset and apex frames. The two components of optical flow images were resized to $28 \times 28$ \emph{pixels} before feeding into our model.

Due to the limited number of samples in the micro-expression datasets, data augmentation was performed during the training to ease over-fitting. A micro-expression involves three key phases: start, peak, and end. Those three phases are labeled as three frames in SAMM and CASME II, i.e., onset, apex and offset frames. We aim to enrich the training data via the specific characteristic of micro-expressions.
In particular, for each micro-expression sequence, we denoted the position of onset, apex and offset frames as ${\textit{Pos}}_{\textit{onset}}$, ${\textit{Pos}}_{\textit{apex}}$ and ${\textit{Pos}}_{\textit{offset}}$, respectively. First, 9 frames in the position of ${\textit{Pos}}_{\textit{onset}} + ({\textit{Pos}}_{\textit{apex}} - {\textit{Pos}}_{\textit{onset}}) \times [0.6, 0.7, 0.8, 0.9]$ and ${\textit{Pos}}_{\textit{apex}} + ({\textit{Pos}}_{\textit{offset}} - {\textit{Pos}}_{\textit{apex}}) \times [0.1, 0.2, 0.3, 0.4, 0.5]$ were selected as the enriched apex frames. By calculating optical flows from the original onset and those enriched apex frames, we gained 9 times of original data for training. Then, all original onset/apex frames and the enriched apex frames obtained by the first steps were rotated between the angles in $[-15^{\circ}, 15^{\circ}]$ with an increment of $5^{\circ}$. Optical flows were computed from the rotated onset and enriched apex frames with the same rotated angle. Performing the two strategies jointly, the original data can be augmented by 70 times.

We adopted stochastic gradient descent with ADAM to learn the network parameters with fixed hyper-parameters to $\beta_1 = 0.9$, $\beta_2 = 0.999$, and $\epsilon = 10^{-8}$. We trained the network in 50 epochs using the mini-batch size of 64 and a learning rate of 0.0005. The loss weight $\lambda = 0.75$ was used. All models were trained on an NVIDIA TITAN X GPU based on the Pytorch deep learning framework. The source code of this work  will be made public after the review period.

\begin{table}
\caption{
Model ablation experiments on modules. Differences and results comparison among the evaluated models on the fold of \emph{SAMM$\rightarrow$CASME II} in the HDE task. \emph{AUdet.}, \emph{AUrel.}, \emph{Fusion} and \emph{DA} denote the AU detection module, the AU relational feature aggregation module, the feature fusion function, and data augmentation, respectively.\emph{ Con.} and \emph{Sum.} mean the feature fusion functions of concatenation and element-wise summation, respectively.}
\label{ma_modules}
\begin{center}
\setlength{\tabcolsep}{0.8mm}{
\begin{tabular}{|m{2.0cm}<{\centering}|c|c|c|c|c|c|c|}
\hline
\multirow{2}*{Method}&\multicolumn{5}{c|}{Differences}&\multicolumn{2}{c|}{Results}\\
\cline{2-8}
&\multirow{2}*{\emph{AUdet.}}&\multirow{2}*{\emph{AUrel.}}&\multicolumn{2}{c|}{\emph{Fusion}}&\multirow{2}*{\emph{DA}}
&\multirow{2}*{WAR}&\multirow{2}*{UAR}\\
\cline{4-5}
&&&\emph{Con.}&\emph{Sum.}&&&\\
\hline
Resnet18&-&-&-&-&\checkmark&0.640&0.532\\
\hline
Dual-Inception (Backbone)&-&-&-&-&\checkmark&0.648&0.544\\
\hline
AUFeat.&\checkmark&-&-&-&\checkmark&0.664&0.560\\
\hline
RelFeat.&\checkmark&\checkmark&-&-&\checkmark&0.676&0.564\\
\hline
ConFeat.&\checkmark&\checkmark&\checkmark&-&\checkmark&0.686&0.579\\
\hline
MER-auGCN w/o DA &\checkmark&\checkmark&-&\checkmark&-&0.659&0.511\\
\hline
MER-auGCN (Ours)&\checkmark&\checkmark&-&\checkmark&\checkmark&0.708&0.595\\
\hline
\end{tabular}}
\end{center}
\end{table}

\begin{table*}
\caption{Model ablation experiments: the effectiveness of Sigmoid attention and balanced detection loss. Differences and results comparison among the evaluated models on the fold of \emph{SAMM$\rightarrow$CASME II} in the HDE task. \emph{SoftAtt.}, \emph{SigAtt.}, \emph{FocLos.} and \emph{BalLos.} denote the Softmax attention strategy, the Sigmoid attention strategy, the original focal loss, and the proposed balanced detection loss, respectively.}
\label{ma_audetec}
\begin{center}
\setlength{\tabcolsep}{1.5mm}{
\begin{tabular}{|c|c|c||c|c|c|c|}
\hline
\multirow{3}*{Method}&\multicolumn{4}{c|}{\multirow{2}*{Differences in AU detection module} }&\multicolumn{2}{c|}{\multirow{2}*{Results}}\\
&\multicolumn{4}{c|}{}&\multicolumn{2}{c|}{}\\
\cline{2-7}
&\emph{SoftAtt.}&\emph{SigAtt.}& \emph{FocLos.}&\emph{BalLos.}&WAR&UAR\\
\hline
SoftmaxAtt.&\checkmark&-&-&\checkmark&0.705&0.566\\
\hline
FocalLos.&&\checkmark&\checkmark&-&0.704&0.582\\
\hline
MER-auGCN (Ours)&-&\checkmark&-&\checkmark&0.708&0.595\\
\hline
\end{tabular}}
\end{center}
\end{table*}

\begin{table}
\caption{Model ablation experiments: the effectiveness of the final AU knowledge-graph. Differences and results comparison among the evaluated models on the fold of \emph{SAMM$\rightarrow$CASME II} in the HDE task. \emph{Densely}, \emph{Original}, and \emph{Final} denote the densely connected graph with the same 12 AU nodes in the final AU knowledge-graph, the original AU knowledge-graph shown in Fig. \ref{graph}(a), and the final AU knowledge-graph used in our proposed model shown in Fig.\ref{graph}(b), respectively.}
\begin{center}
\label{ma_aurel}
\setlength{\tabcolsep}{1.8mm}{
\begin{tabular}{|m{2.0cm}<{\centering}|c|c|c|c|c|}
\hline
\multirow{3}*{Method}&\multicolumn{3}{c|}{Differences in AU relational}&\multicolumn{2}{c|}{\multirow{2}*{Results}}\\
&\multicolumn{3}{c|}{feature aggregation module}&\multicolumn{2}{c|}{}\\
\cline{2-6}
&\emph{Densely}&\emph{Original}&\emph{Final}&WAR&UAR\\
\hline
DenselyG.&\checkmark&-&-&0.699&0.568\\
\hline
OriginalG.&-&\checkmark&-&0.690&0.569\\
\hline
MER-auGCN (Ours)&-&-&\checkmark&0.708&0.595\\
\hline
\end{tabular}}
\end{center}
\end{table}
\subsection{Model ablation}
In this section, we performed detailed ablation studies on the \emph{SAMM$\rightarrow$CASME II} sub-task in the HDE task from two aspects: 1) Model ablation experiments on modules. To select the most suitable model components in MER-auGCN and to verify the effectiveness of our data augmentation strategy, we designed several baseline methods to analyze our MER-auGCN model and show the benefits of each module;  2) Model ablation experiments on strategies used in modules. Going deeper into the strategies used in the modules, we first implemented the ablation experiments on the Softmax attention and the original focal loss \cite{Lin2017a} to verify the effectiveness of the Sigmoid attention and the improved focal loss (balanced detection loss) in the AU detection module. Then we compared the final AU knowledge-graph with a densely connected graph and the original AU graph showed in Fig.\ref{graph}(a) to demonstrate the usefulness of the final AU knowledge-graph used in the AU relational feature aggregation module.

\subsubsection{Model ablation experiments on modules}
The performance comparison between the baseline methods with different modules and MER-auGCN are shown in Table \ref{ma_modules} with detailed described below.
\begin{itemize}
\item Deep depth network of Resnet18 (\textbf{Resnet18}): We investigated the deep depth network Resnet18 \cite{He2016a} as a backbone candidate to compare its performance with that of the shallow network: Dual-Inception \cite{Zhou2019}. In order to feed the two components of the optical flow into Resnet18, we expanded the basic Renset18 to a two-path network, and each path contained four basic blocks of Resnet18. Feature maps after the two-path blocks were then concatenated for prediction. The two components of the optical flow images were cropped to $112 \times 112$ \emph{pixels} before feeding into Resnet18. WAR and UAR in Table \ref{ma_modules} reveal that Dual-Inception works slightly better. Thus, we adopted Dual-Inception as our backbone model.

\item AU feature learning (\textbf{AUFeat.}): Compared with the backbone of Dual-Inception, this baseline added the AU detection module to learn AU features and then performed MER on the representation $z_o$ (the sum-pooled representation of the AU features). AUFeat. outperformed Dual-Inception as shown in Table \ref{ma_modules}. This result verified the effectiveness of our AU detection module with the attention mechanism and the balanced detection loss.

\item AU relational feature aggregation (\textbf{RelFeat.}): With the guidance of the AU knowledge-graph, RelFeat. leveraged the GCN to aggregate the AU-level features into a micro-expression-level feature representation ($z_r$) for MER. The superior performance of RelFeat. to AUFeat. indicates that in objective class-based MER, it is helpful to construct the AU knowledge-graph of micro-expressions and update the relational feature of each AU by aggregating from its neighbor nodes accordingly.

\item Fusing features by concatenation (\textbf{ConFeat.}): Different from MER-auGCN which used the element-wise summation to fuse the sum-pooled representation of the AU features $z_o$ and the micro-expression-level representation $z_r$, ConFeat. employed the concatenation function for feature fusion. Table \ref{ma_modules} shows that MER-auGCN achieved a better performance. Thus, we fused the two representations via summation in the remaining of our experiments.

\item MER-auGCN without data augmentation (\textbf{MER-auGCN w/o DA}): Compared with MER-auGCN w/o DA, the better performance of MER-auGCN shows that our data augmentation strategy is highly effective.
\end{itemize}
\begin{table*}
\caption{Performance comparison with the state-of-the-art objective class-based MER methods. The best results are highlighted in bold.} \label{all_results}
\subtable[Performance comparison on the HDE task.]{
\label{hde_result}
\setlength{\tabcolsep}{0.7mm}{
\begin{tabular}{|m{2.3cm}<{\centering}|p{1.4cm}<{\centering}|p{1.4cm}<{\centering}|p{1.4cm}<{\centering}|p{1.4cm}<{\centering}|p{1.0cm}<{\centering}|p{1.0cm}<{\centering}|}
\hline
\multirow{2}*{Method}&\multicolumn{2}{c|}{CASME II$\rightarrow$SAMM}&\multicolumn{2}{c|}{SAMM$\rightarrow$CASME II}&\multicolumn{2}{c|}{Avg.}\\
\cline{2-7}
&WAR&UAR&WAR&UAR&WAR&UAR\\
\hline
LBP-TOP \cite{Zhao2007}&0.338&0.327&0.232&0.316&0.285&0.322\\
\hline
3DHOG \cite{Polikovsky2009}&0.353&0.269&0.373&0.187&0.363&0.228\\
\hline
HOOF \cite{Liu2016}&0.441&0.349&0.265&0.346&0.353&0.348\\
\hline
Khor et al.\cite{Khor2018}&0.485&0.382&0.384&0.322&0.435&0.352\\
\hline
Peng et al.\cite{Peng2018}&0.544&0.440&0.578&0.337&0.561&0.389\\
\hline
\textbf{MER-auGCN (Ours)}&\textbf{0.662}&\textbf{0.588}&\textbf{0.708}&\textbf{0.595}&\textbf{0.685}&\textbf{0.592}\\
\hline
\end{tabular}}
\label{hde_result}
}
\subtable[Performance comparison on the CDE task.]{
\label{cde_result}
\setlength{\tabcolsep}{0.9mm}{
\begin{tabular}{|m{3.0cm}<{\centering}|m{1.0cm}<{\centering}|m{1.0cm}<{\centering}|}
\hline
Method&F1&WF1\\
\hline
LBP-TOP \cite{Zhao2007}&0.400&0.524\\
\hline
3DHOG \cite{Polikovsky2009}&0.271&0.436\\
\hline
HOOF \cite{Liu2016}&0.404&0.527\\
\hline
Knor et al. \cite{Khor2018}&0.393&0.523\\
\hline
 Davison et al.\cite{Davison2018a}&0.454&0.579\\
\hline
Peng et al.\cite{Peng2018}&0.639&0.733\\
\hline
\textbf{MER-auGCN (Ours)}&\textbf{0.685}&\textbf{0.742}\\
\hline
\end{tabular}}
\label{cde_result}
}
\end{table*}

\begin{table*}
\caption{F1 (\%) comparison with different AU detection methods on the HDE protocol.  The highest score is highlighted in bold.}
\label{aus_result}
\begin{center}
\setlength{\tabcolsep}{1.0mm}{
\begin{tabular}{|c|p{42pt}<{\centering}|p{44pt}<{\centering}|m{55pt}<{\centering}
|p{42pt}<{\centering}|p{44pt}<{\centering}|m{55pt}<{\centering}
|p{42pt}<{\centering}|p{44pt}<{\centering}|m{55pt}<{\centering}|}
\hline
&\multicolumn{3}{c|}{CASME II$\rightarrow$SAMM}&\multicolumn{3}{c|}{SAMM$\rightarrow$CASME II}&\multicolumn{3}{c|}{Avg.}\\
\hline
AU&LBP-TOP\cite{Zhao2007}&STAPF\cite{Li2019}&\textbf{MER-auGCN (Ours)}&LBP-TOP\cite{Zhao2007}&STAPF\cite{Li2019}&\textbf{MER-auGCN (Ours)}& LBP-TOP\cite{Zhao2007} &STAPF\cite{Li2019}&\textbf{MER-auGCN (Ours)}\\
\hline
1&0.00&\textbf{60.00}&46.36&0.00&50.00&\textbf{71.72}&0.00&55.00&\textbf{59.04}\\
\hline
2&0.00&59.57&\textbf{76.58}&0.00&41.67&\textbf{69.49}&0.00&50.62&\textbf{73.04}\\
\hline
4&44.00&54.55&\textbf{77.48}&0.00&41.67&\textbf{94.62}&22.00&48.11&\textbf{86.05}\\
\hline
9&0.00&\textbf{54.55}&36.31&0.00&\textbf{66.67}&26.32&0.00&\textbf{60.61}&31.32\\
\hline
12&0.00&45.28&\textbf{55.98}&0.00&53.84&\textbf{58.56}&0.00&49.56&\textbf{57.27}\\
\hline
14&0.00&\textbf{64.86}&54.35&0.00&6\textbf{2.50}&46.28&0.00&\textbf{63.68}&50.32\\
\hline
17&0.00&\textbf{72.72}&53.51&0.00&58.82&\textbf{60.72}&0.00&\textbf{65.77}&57.12\\
\hline
Avg.&6.29&\textbf{58.79}&57.22&0.00&53.60&\textbf{61.10}&3.15&56.20&\textbf{59.16}\\
\hline
\end{tabular}}
\end{center}
\end{table*}

\subsubsection{Model ablation experiments on strategies used in modules}
The performance comparison between the baselines with different strategies in modules and MER-auGCN are shown in Table \ref{ma_audetec} and Table \ref{ma_aurel} with detailed described below.

\begin{itemize}

\item \textbf{Effectiveness of the Sigmoid attention}: We conducted a group of experiment (i.e., \textbf{SoftAtt.} vs MER-auGCN) to verify the effectiveness of the Sigmoid attention strategy used in the AU detection module. As displayed in Table \ref{ma_audetec}, compared with SoftAtt. which applies Softmax attention mechanism for AU feature learning, MER-auGCN achieves a higher WAR and UAR with the help of the Sigmoid attention mechanism. It is because in the AU feature learning, the characteristics of each element of a feature are not necessarily mutually exclusive, the Sigmoid attention allows for greater flexibility, i.e., allowing multiple elements in one feature to get larger weights at the same time.

\item \textbf{Effectiveness of the balanced detection loss}: Table \ref{ma_audetec} illustrates the performance comparison among the baseline of \textbf{FocalLos.} and MER-auGCN. It is noteworthy that MER-auGCN which leverages our proposed balanced detection loss outperforms the FocalLos. that uses the original focal loss. It validates the effectiveness of our proposed balanced detection loss and indicates that adjusting the typically focal loss to the specific applications with imbalanced data may help to improve the model performance.

\item \textbf{Effectiveness of the final AU knowledge-graph}: To verify the effectiveness of the final AU knowledge-graph used in MER-auGCN, we compared the performance of our proposed method with the baselines which aggregating AU features in GCN by the guidance of a densely connected graph with the same 12 AU nodes in the final AU knowledge-graph (\textbf{DenselyG.}), and by the guidance of the original AU knowledge-graph (\textbf{OriginalG.}) shown in Fig. \ref{graph}(a), respectively. As illustrated in Table \ref{ma_aurel}, we can clearly see the effectiveness. It is noted that MER-auGCN achieves 0.90\% and 2.70\% performance boost w.r.t WAR and UAR when compared with DenselyG.. Similar performance improvements can be found in Table \ref{ma_aurel} where MER-auGCN outperforms OriginalG. with the improvement of 1.80\% and 2.60\% in terms of WAR and UAR, respectively. These comparisons validate the reasonableness of our constructed AU knowledge-graph, which also further verify the reasonableness of the relationship between action unit and objective classes I-V shown in Table \ref{objective_class}.

\end{itemize}

\subsection{Comparison with the state-of-the-arts of MER}
In this section, we compare our MER-auGCN model with the state-of-the-art objective class-based MER methods. The full results are provided in Table \ref{all_results}.

\subsubsection{HDE task}
Table \ref{hde_result} shows the comparison with prior results reported in the literature on the HDE task for objective class-based MER. Our method surpasses all existing methods by a good margin and establishes the new state-of-the-art. Our model utilizes the same feature of TV-L1 optical flow as \cite{Khor2018} and outperforms it by about 0.25 on WAR and UAR, which clearly demonstrates the advantage of MER feature learning through simultaneous AU detection and graph-based feature aggregation. We also achieved better performance than the transferring learning method \cite{Peng2018}, mostly because we explicitly learn the AU features and aggregate the AU-level features into the micro-expression-level feature with the input of optical flow other than the apex frame. This is also the reason why so many researchers of MER adopted optical flow as the input of models, especially in cross-database recognition tasks: optical flow focuses on extracting the motion feature of samples, which suppresses the facial identity amongst different databases to a certain degree.  Meanwhile, our method outperforms the handcrafted feature based methods \cite{Zhao2007, Polikovsky2009, Liu2016, Davison2018a}, which indicates that deep learning methods has promising prospects in MER with small databases when equipped with a proper feature learning model and data augmentation strategy.

\subsubsection{CDE task}
We further evaluate our proposed model on the CDE task. The results and comparison with previous methods are listed in Table \ref{cde_result}. Our model again achieved the state-of-the-art performance (F1: 0.685, WF1: 0.742). This outstanding performance shows the effectiveness and generality of our proposed MER-auGCN for learning effective global micro-expression-level representation by leveraging the interactions between local AU features.

\subsection{AU detection on the HDE task}
The last experiment was conducted to evaluate the performance of our model for micro-expression AU detection. We compared our model with STAPF, a micro-expression AU detection method recently proposed in \cite{Li2019}, based on the F1 score. As STAPF did not perform AU detection with the CDE protocol, we only compared the results using the HDE protocol. For a fair comparison, we used the same data set as STAPF, which contains 228 samples of CASME II and 78 samples of SAMM, all with 7 AUs (AU1, AU2, AU4, AU9, AU12, AU14, and AU17).

    The results of our empirical evaluation are summarized in Table \ref{aus_result}. Clearly, MER-auGCN obtained very competitive performance on the fold of \emph{CASME II$\rightarrow$ SAMM} and got superior performance on the fold of \emph{SAMM$\rightarrow$ CASME II} when compared with STAPF. When averaged over all seven AUs, it got an F1 score of 59.16\% and outperformed STAPF (56.20 \%). Note that MER-auGCN achieved these AU detection results with \textbf{a single, unified model} while STAPF trains a different model for each AU and needs seven models in total. Clearly, our model can be easily scaled when more AUs are included in real-world scenarios. It is also worth mentioning that STAPF features have to be learned separately and prior to feature aggregation for MER while ours is an \textbf{end-to-end model} through the joint optimization of the AU detection and MER losses. Through this experiment, it is clear that the AU detection module in MER-auGCN can effectively learn the AU features, which lays the foundation for accurate MER and other potential applications of AU-based facial analysis.

\section{Conclusion}
In this paper, we proposed a novel model, MER-auGCN, for simultaneous AU detection and feature aggregation in objective class-based MER. In MER-auGCN, features for all the AUs are learned in a unified model, eliminating the error-prune landmark detection process and tedious separate training for each AU. We also introduced an AU knowledge-graph to capture the relationship between AUs for feature aggregation. Experiments on two tasks in MEGC2018 showed that MER-auGCN significantly outperformed the current state-of-the-arts in MER. Additionally, our micro-expression AU detection results showed that MER-auGCN can learn effective AU features with a unified model.

\section{Acknowledgements}
\textcolor{black}{
This work was supported in part by the National Natural Science Foundation of China under Grant 61672267, Grant U1836220, in part by the Postgraduate Research \& Practice Innovation Program of Jiangsu Province under Grant KYCX19\_1616, Qing Lan Talent Program of Jiangsu Province, and Jiangsu Engineering Research Center of big data ubiquitous perception and intelligent agriculture applications.}

\bibliographystyle{IEEEtran}
\bibliography{ref_new}
\end{document}